\documentclass{article}

\usepackage{float}
\usepackage{microtype}
\usepackage{graphicx}
\usepackage{subfigure}
\usepackage{booktabs} % for professional tables

\usepackage{hyperref}

\usepackage[accepted]{icml2024}

\usepackage{amsmath}
\usepackage{amssymb}
\usepackage{mathtools}
\usepackage{amsthm}

\usepackage[capitalize,noabbrev]{cleveref}

\usepackage{hyperref}
\usepackage{url}
\usepackage{amsfonts}
\usepackage{textcomp}
\usepackage{amssymb}
\usepackage{relsize}

\usepackage{float}
\usepackage{multirow}
\usepackage{array}
\usepackage{caption}
\usepackage{subcaption}
\usepackage[english]{babel}
\usepackage{amsthm}

\theoremstyle{plain}

\theoremstyle{definition}

\theoremstyle{remark}

\usepackage[textsize=tiny]{todonotes}

\icmltitlerunning{Perceiver-based CDF Modeling for Time Series Forecasting}

\begin{document}

\twocolumn[
\icmltitle{Perceiver-based CDF Modeling for Time Series Forecasting}



\icmlsetsymbol{equal}{*}

\begin{icmlauthorlist}
\icmlauthor{Cat P. Le}{equal,sch}
\icmlauthor{Chris Cannella}{equal,sch}
\icmlauthor{Ali Hasan}{sch}
\icmlauthor{Yuting Ng}{sch}
\icmlauthor{Vahid Tarokh}{sch}
\end{icmlauthorlist}

\icmlaffiliation{sch}{Department of Electrical and Computer Engineering, Duke University}

\icmlcorrespondingauthor{Cat P. Le}{cat.le@duke.edu}

\icmlkeywords{Machine Learning, ICML}

\vskip 0.3in
]





\begin{abstract}
Transformers have demonstrated remarkable efficacy in forecasting time series data. However, their extensive dependence on self-attention mechanisms demands significant computational resources, thereby limiting their practical applicability across diverse tasks, especially in multimodal problems. In this work, we propose a new architecture, called perceiver-CDF, for modeling cumulative distribution functions (CDF) of time series data. Our approach combines the perceiver architecture with a copula-based attention mechanism tailored for multimodal time series prediction. By leveraging the perceiver, our model efficiently transforms high-dimensional and multimodal data into a compact latent space, thereby significantly reducing computational demands. Subsequently, we implement a copula-based attention mechanism to construct the joint distribution of missing data for prediction. Further, we propose an output variance testing mechanism to effectively mitigate error propagation during prediction. To enhance efficiency and reduce complexity, we introduce midpoint inference for the local attention mechanism. This enables the model to efficiently capture dependencies within nearby imputed samples without considering all previous samples. The experiments on the unimodal and multimodal benchmarks consistently demonstrate a 20\% improvement over state-of-the-art methods while utilizing less than half of the computational resources.
\end{abstract}

\section{Introduction}
Time-series prediction remains an enduring challenge since it requires effectively capturing global patterns (e.g., annual trends) and localized details (e.g., abrupt disruptions). This challenge becomes particularly pronounced when dealing with non-synchronized, incomplete, high-dimensional, or multimodal input data. Consider a time series consisting of $N$ regularly-sampled and synchronously-measured values, where measurements are taken at intervals of length $T$. If the time-step is unobserved at rate $r$, then there are $(1-r)NT$ observed values that are relevant for inference. Consider an asynchronously-measured time series, where input variables are observed at different times, resulting in each time-step having only $1/N$ of its variables observed. In this scenario, only $(1-r)T$ values remain relevant for inference within the time series. Consequently, employing a synchronous model to address non-synchronized time series results in a missingness rate of $(N-1)/N$. This missingness rate grows rapidly as the number of variables increases, reaching 95\% with just $20$ variables in the time series. When designing an architecture to handle missing data, it is crucial to utilize techniques for approximating missing values while ensuring the computational overhead does not exceed the effort required to extract valuable insights from the observed data. To this end, a transformer model~\citep{drouin2022tactis} with attention-based mechanism~\citep{vaswani2017attention} is tailored for time series. This model tokenizes input variables and utilizes a transformer-based encoding and decoding approach, making it a suitable choice for modeling non-synchronized time series data. Tokenization also offers significant advantages for missing data, as unobserved data can be seamlessly excluded from the token stream. % and using copulas allow easy marginalization of excluded items
Additionally, this model utilizes a copula-based structure~\citep{nelsen2006introduction} to represent the sequence distribution to further enhance the prediction performance. 
Particularly, it learns the joint distribution with a non-parametric copula, which is a product of conditional probabilities. To ensure that the product results in a valid copula, it considers permutations of the margins during training such that a level of permutation invariance occurs. However, this process yields an exchangeable class of copulas in the limit of infinite permutations, diminishing the utility of the non-parametric copula. Furthermore, the transformer architecture poses significant computational demands related to the self-attention mechanism.

In this paper, we introduce the perceiver-CDF model, which utilizes the perceiver architecture~\citep{jaegle2021perceiverio} and the attention-based copulas~\citep{nelsen2006introduction}, to enhance multimodal time series modeling and address computational efficiency challenges. Particularly, our model consists of the perceiver-based encoder and the copula-based decoder, enabling the incorporation of a more general class of copulas that are not exchangeable. The class of copulas used in our approach are the \emph{factor copulas}, which are conditionally exchangeable based on the factor. Initially, the perceiver-CDF model transforms the input variables into temporal embeddings through a combination of input embedding and positional encoding procedures. In this phase, the observed and the missing data points are encoded, with the value of missing data points masked. Subsequently, our proposed model utilizes a \emph{latent} attention-based mechanism~\citep{jaegle2021perceiverio} from the perceiver to efficiently map the input embeddings to a lower-dimensional latent space. Since all subsequent computations are performed within this compact latent space, it helps reduce the complexity from a quadratic to a sub-quadratic level. Lastly, the copula-based decoder formulates the joint distribution of missing data using latent embeddings. This distribution undergoes a sampling process to yield the predicted outcomes. Our model can effectively handle synchronized, non-synchronized, and multimodal data, expanding its applicability to diverse domains.

To demonstrate the efficacy of our proposed model, we conduct extensive experiments on the unimodal datasets from the Monash Time Series Forecasting Repository~\citep{godahewa2021monash} (i.e., electricity, traffic, and fred-md) and the multimodal datasets, such as room occupation~\citep{misc_occupancy_detection__357}, interstate traffic~\citep{misc_metro_interstate_traffic_volume_492}, and air quality~\citep{misc_beijing_multi-site_air-quality_data_501}. We also conduct memory consumption scaling experiments using random walk data to demonstrate the memory efficiency of our approach. The results demonstrate the competitive performance of our model compared to the state-of-the-art methods, including TACTiS~\citep{drouin2022tactis}, GPVar~\citep{salinas2019high}, SSAE-LSTM~\citep{zhu2021multimodal}, and AR~\citep{kalliovirta2015gaussian} while utilizing as little as 10\% of the computational resources. 

%The contributions of our paper are summarized below:

%\begin{itemize}
%    \item We propose a perceiving copulas architecture, which incorporates the latent-attention mechanism with the attentional-copula structure for multimodal time series forecasting.
%    \item We introduce midpoint inference for the local attention mechanism to capture the dependencies between imputed samples while reducing computational complexity.
%    \item We present the output variance test mechanism to mitigate error propagation, thereby significantly enhancing prediction performance
%\end{itemize}

%-------------------------------------------------------
\section{Related Work}

%%% TODO: Copula for time-series. Usually only applied per time marginal. Across time, I think they are only some variant of Gaussian processes.

%\subsection*{Related Works}
%\textbf{\normalsize Related Works.} 
Neural networks for time series forecasting~\citep{zhang1998forecasting} have undergone extensive research and delivered impressive results when compared to classical statistical methods~\citep{box2015time, hyndman2008forecasting, yanchenko2020stanza}. Notably, both convolutional~\citep{chen2020probabilistic} and recurrent neural networks~\citep{connor1994recurrent, shih2019temporal,hochreiter1997long} have demonstrated the power of deep neural networks in learning historical patterns and leveraging this knowledge for precise predictions of future data points. Subsequently, various deep learning techniques have been proposed to address the modeling of regularly-sampled time series data~\citep{oreshkin2019n, le2020probabilistic, de2020normalizing, lim2021time, benidis2022deep}. Most recently, the transformer architecture, initially designed for sequence modeling tasks, has been adopted extensively for time series forecasting~\citep{li2019enhancing, lim2021temporal, muller2021transformers}. Using the properties of the attention mechanism, these models excel at capturing long-term dependencies within the data, achieving remarkable results. In addition to these developments, score-based diffusion models~\citep{tashiro2021csdi} achieved competitive performance in forecasting tasks. However, it is worth noting that the majority of these approaches are tailored for handling regularly sampled and synchronized time series data. Consequently, they may not be optimal when applied to non-synchronized datasets.
In financial forecasting, the copula emerges as a formidable tool for estimating multivariate distributions~\citep{aas2009pair, patton2012review, krupskii2020flexible, grosser2022copulae, mayer2023estimation}. Its computational efficiency has led to its use in the domain adaptation contexts~\citep{lopez2012semi}. Moreover, the copula structure has found utility in time series prediction when coupled with neural architectures like LSTMs~\citep{lopez2012semi} and the transformer~\citep{drouin2022tactis}, enabling the modeling of irregularly sampled time series data. While previous research has explored non-synchronized methods~\citep{chapados2007augmented, shukla2021multi}, their practicality often falters due to computational challenges. For instance, the transformer architecture with copulas~\citep{drouin2022tactis} is proposed and applicable to both synchronized and non-synchronized datasets. Nonetheless, the inherent computational overhead associated with the self-attention mechanism poses limitations, particularly when applied to high-dimensional inputs such as multimodal data. To mitigate the computational complexity, we propose the perceiver-CDF model which utilizes the perceiver architecture~\citep{jaegle2021perceiver, jaegle2021perceiverio} as the encoder, paired with a copula-based decoder. We also utilize the midpoint inference~\citep{liu2019naomi} for the local inference during the decoding phase of the model. This approach restricts conditioning and effectively embodies a form of sparse attention~\citep{child2019generating, tay2020sparse, roy2021efficient}, although the sparsity pattern is determined through a gap-filling process.

%--------------------------------------------
\begin{figure*}[t]
\begin{center}
\centering
\includegraphics[width=0.85\textwidth]{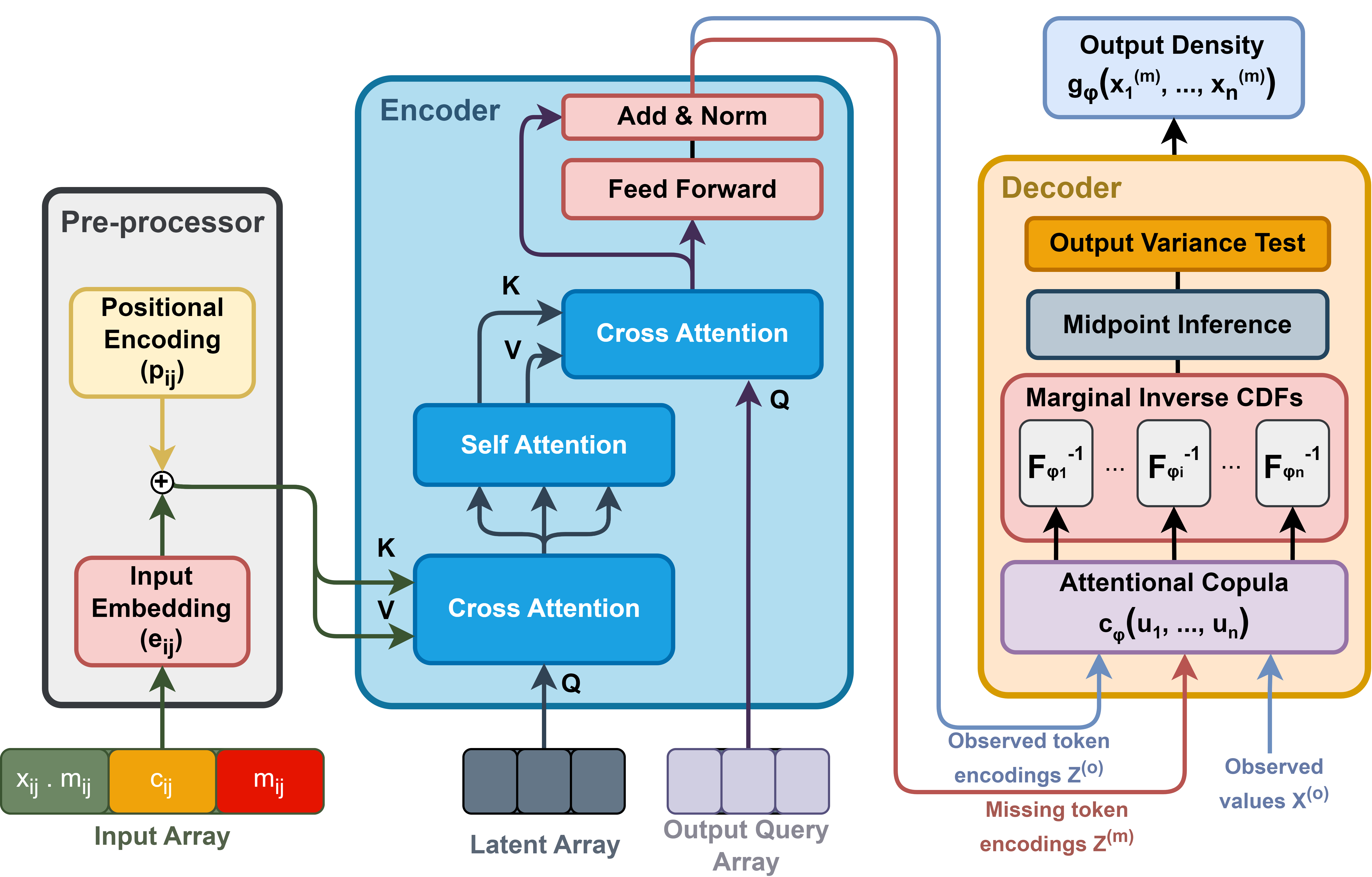}
\caption{The overview architecture of perceiver-CDF model. The pre-processor includes input embedding, and positional encoding layers to capture temporal dependencies in the input data. The encoder uses the cross-attention mechanisms to map the embedding to a lower-dimensional latent space. The decoder constructs the joint distribution of missing data using the copula-based structure.}
\label{fig:practis}
\end{center}
\end{figure*}
%--------------------------------------------

%-------------------------------------------------------------------
\section{Perceiver-CDF}

In this study, we introduce a new perceiver-CDF model designed for multimodal time series prediction. 
The model represents a groundbreaking advancement in generative modeling for time series data and comprises three key components: the pre-processor, the encoder, and the decoder. Drawing inspiration from transformers and other transformer-based models~\citep{drouin2022tactis}, the pre-processor utilizes the self-attention mechanism~\citep{vaswani2017attention} to encode input time series variables, effectively transforming them into a sequence of generalized tokens. This transformation enables the model to process and analyze the temporal aspects of the data efficiently. Following this, the encoder, based on the perceiver model~\citep{jaegle2021perceiver, jaegle2021perceiverio}, applies cross-attention to the resulting token sequence to generate a conditional distribution of inferred variables using a parameterized copula. Particularly, it converts the complex input tokens into a compact latent space. This step is crucial for computational efficiency, as subsequent computations operate within this lower-dimensional space. 
Subsequently, the copula-based decoder is utilized to learn the joint distribution of missing data and observed data, facilitating future predictions. During the training process, we implement midpoint inference in the decoder for local inference instead of random imputation, contributing to further computational reduction. This mechanism also aids in establishing dependencies between nearby imputed samples. To validate and refine predictions, we introduce a variance testing mechanism for output prediction. If the prediction results exhibit instability and the variance exceeds a predefined threshold, the imputation is deemed unreliable for future predictions. It will be masked to prevent error propagation. 

The proposed model utilizes the advantages of both the self-attention mechanism and latent-variable-based attention mechanisms from perceivers. Notably, it enables the modeling of dependencies between covariates, which can converge into a factor copula~\citep{oh2017modeling, krupskii2013factor} described as follows:
\begin{multline}
    C(u_1,\cdots,u_d) = \\
    \int_{[0,1]^k}\prod_{j=1}^d F_{j|Z_1,\cdots,Z_k}(u_j|z_1,\cdots,z_k)dz_1\cdots z_k
\end{multline} 

The factor copula emerges as an optimal choice for modeling high-dimensional data, Particularly, it permits the specification of a parametric form with linear $\mathcal{O}(n)$ dependency parameters, rather than the computationally burdensome $\mathcal{O}(n^2)$ parameters, where $n$ is the number of observed variables. Furthermore, the factor copula model proves invaluable in scenarios where the interdependence among observed variables is contingent upon a limited number of unobserved variables, particularly in situations where there is a presence of tail asymmetry or tail dependence within the dataset. Across various multivariate contexts, the reliance on observed variables can be explained through latent variables. Crucially, this approach dispenses with the assumption of exchangeability, allowing the copula to adopt a more general form. The perceiver considers this structure conditioned on latent variables. As a result, our proposed model is capable of effectively handling multimodal input data, while significantly reducing computational complexity. 
Overall, the perceiver serves as the backbone of our model, enabling it to efficiently process a wide spectrum of data types. The overview architecture of our proposed perceiver-CDF model is illustrated in Figure~\ref{fig:practis}. Subsequently, we delve into the intricate components of this model to provide a comprehensive understanding.

\subsection{Perceiver-based Encoder}
Let $\mathcal{X}$ denote the a time series of interest, with $\mathcal{X} = \{X_{1}, X_{2}, \ldots, X_{i}, \ldots\}$. Each element $X_{i}$ is defined as a quadruple: $X_{i} = (v_{i}, c_{i}, t_{i}, m_{i})$, where $v_{i}$ is the value, $c_{i}$ is an index identifying the variable, $t_{i}$ is a time stamp, and $m_{i}$ is a mask indicating whether the data point is observed or needs to be inferred (i.e., missing data). For synchronously measured time series, we can organize it into a data matrix denoted as $X_{c, t}$. This matrix has rows corresponding to individual variables and columns corresponding to different timestamps when measurements were recorded. 
First, the pre-processor generates embeddings for each data point, $\vec{x}_{i}$, which includes the value $v_{i}$, a learned encoding for the variable $c_{i}$, an additive sinusoidal positional encoding indicating the position of $t_{i}$ within the overall time series, and the mask $m_{i}$. 
Subsequently, these embeddings are passed through the perceiver-based encoder. Here, the encoder leverages a predefined set of learned latent vectors $\vec{u}_{L}$ for the cross-attention mechanism, denoted by $\mathcal{A_C}(K, Q, V)$, where $K$ is a set of keys, $Q$ is a query, and $V$ is a set of values. Through the utilization of learned key and value-generating functions, $K_\mathtt{latent}()$ and $V_\mathtt{latent}()$, the encoder derives latent vectors $\vec{w}_{L}$, which effectively encapsulate the temporal information through cross-attention with the set of observed vectors $\vec{X}_{O}$ as follows:
\begin{equation}
\vec{w}_{L} = \mathcal{A_C} \left (K_\mathtt{latent}(\vec{X}_{O}), \vec{u}_{L}, V_\mathtt{latent}(\vec{X}_{O}) \right )
\end{equation}

%By utilizing learned functions for generating keys and values, $\mathtt{key\_encode}()$ and $\mathtt{value\_encode}()$, TACTiS derives a tokenized representation, denoted as $\vec{z}_{i}$, for each data point. This is achieved by passing the input embeddings through a stack of residual layers, as follows:
%\begin{equation*}
%\vec{z}_{i} = \mathtt{self\_attention}(\mathtt{key\_encode}(\vec{x}_{\neg i}), \vec{x}_{i}, \mathtt{value\_encode}(\vec{x}_{\neg i}))
%\end{equation*}

Following additional self-attention-based processing on the latent vectors, the perceiver-based encoder proceeds to employ cross-attention with the latent vector set $\vec{W}$, to generate tokens for each data point. This operation involves using the learned key-generating function $K_\mathtt{encode}()$, the query-generating function $Q_\mathtt{encode}()$, and the value-generating functions $V_\mathtt{encode}()$, to derive token vectors $\vec{z}_{i}$ as follows:
\begin{equation}
\vec{z}_{i} = \mathcal{A_C} \left (K_\mathtt{encode}(\vec{W}), Q_\mathtt{encode}(\vec{x}_{i}), V_\mathtt{encode}(\vec{W}) \right )
\end{equation}

Aligned with the perceiver architecture, the number of latent features $L$ is intentionally maintained at a considerably smaller scale compared to the total number of data points $N$. This strategic choice serves to manage computational complexity, which scales at $\mathcal{O}(NL)$. The initial cross-attention step in our model assumes a pivotal role by encoding a comprehensive global summary of the observed data from the time series into a set of concise latent vectors. These latent vectors effectively capture the essential information embedded within the entire dataset. Subsequently, our proposed perceiver-CDF model generates tokens for each individual data point by efficiently querying relevant global information from the previously obtained latent summary in the second cross-attention step. This process ensures that each token encompasses vital contextual details drawn from the overall dataset, as necessitated.

%-----
\subsection{Copula-based Decoder}
Next, the decoder is specifically designed to learn the joint distribution of the missing data points conditioned on the observed ones.  To achieve this, the attention-based decoder is trained to mimic a non-parametric copula~\citep{nelsen2006introduction}. Let $x^{(m)}$ and $x^{(o)}$ represent the missing and observed data points, respectively. Let $F_i$ be the $i^\text{th}$ marginal cumulative distribution function (CDF) and $f_i$ be the marginal probability density function (PDF). The copulas, under Sklar's theorem~\citep{sklar1959fonctions}, allow for separate modeling of the joint distribution and the marginals, which has particular relevance to the case of sequence modeling.
Similar to TACTiS~\citep{drouin2022tactis}, we employ a normalizing flow technique known as Deep Sigmoidal Flow~\citep{huang2018neural} to model the marginal CDF. The marginal PDF is obtained by differentiating the marginal CDF. The copula-based structure $g_\phi$ is described as follows:
\begin{multline}
g_\phi \left(X^{(m)}\right) =
c_{\phi_c} \left (F_{\phi_1} \left(x^{(m)}_1\right), \ldots, F_{\phi_{n_m}}\left(x^{(m)}_{n_m}\right) \right) \\ 
\times f_{\phi_1} \left(x^{(m)}_1\right) \times \ldots \times f_{\phi_{n_m}} \left(x^{(m)}_{n_m}\right),
\end{multline}

where $X^{(m)} = \{x^{(m)}_1,\ldots, x^{(m)}_{n_m}\}$ and $c_{\phi_c}$ is the density of a copula, $c_{\phi_c} (F_{\phi_1} (x^{(m)}_1), \ldots, F_{\phi_{n_m}}(x^{(m)}_{n_m})) = c_{\phi_{c1}} (F_{\phi_1} (x^{(m)}_1)) \times c_{\phi_{c2}} (F_{\phi_2} (x^{(m)}_2) \big| F_{\phi_1} (x^{(m)}_1)) \times \ldots \times c_{\phi_{cn_m}}  (F_{\phi_{n_m}} (x^{(m)}_{n_m}) \big| F_{\phi_1} (x^{(m)}_1), \ldots,  F_{\phi_{n_m-1}} (x^{(m)}_{n_m-1}))$.

%described as follows:
%\begin{equation*}
%\begin{aligned}
%& c_{\phi_c} \left (F_{\phi_1} \left(x^{(m)}_1\right), \ldots, F_{\phi_{n_m}}\left(x^{(m)}_{n_m}\right) \right)\\
%& = c_{\phi_{c1}} \left(F_{\phi_1} \left(x^{(m)}_1\right) \right) \times \ldots \times c_{\phi_{cn_m}} \left(F_{\phi_{n_m}} \left(x^{(m)}_{n_m}\right) \Big| F_{\phi_1} \left(x^{(m)}_1\right), \ldots,  F_{\phi_{n_m-1}} \left(x^{(m)}_{n_m-1} \right) \right)
%\end{aligned}
%\end{equation*}

%\begin{equation*}
%\begin{aligned}
%& c_{\phi_c} \left (F_{\phi_1} \left(x^{(m)}_1\right), \ldots, F_{\phi_{n_m}}\left(x^{(m)}_{n_m}\right) \right)\\
%& = c_{\phi_{c1}} \left(F_{\phi_1} \left(x^{(m)}_1\right) \right) \times c_{\phi_{c2}} \left(F_{\phi_2} \left(x^{(m)}_2\right) \Big| F_{\phi_1} \left(x^{(m)}_1\right) \right) \times \ldots \times \\
%& \quad c_{\phi_{cn_m}} \left(F_{\phi_{n_m}} \left(x^{(m)}_{n_m}\right) \Big| F_{\phi_1} \left(x^{(m)}_1\right), \ldots,  F_{\phi_{n_m-1}} \left(x^{(m)}_{n_m-1} \right) \right)
%\end{aligned}
%\end{equation*}
    
During the decoding phase, our model selects a permutation, denoted as $\gamma$, from all data points, ensuring that observed data points come before those awaiting inference. The decoder then utilizes the self-attention mechanism $\mathcal{A_S}(K,Q,V)$ with the learned key and value functions, $K_\mathtt{decode}()$ and $V_\mathtt{decode}()$, to derive distributional parameters, $\theta_{\gamma(i)}$, for each data point awaiting inference as follows:
\begin{multline}
    \theta_{\gamma(i)} = \mathcal{A_S} \Big( K_\mathtt{decode} \left (\vec{z}_{\gamma(j)} \right), \vec{z}_{\gamma(i)}, V_\mathtt{decode} \left( \vec{z}_{\gamma(j)} \right) \Big)
\end{multline}

where ${\gamma(j) < \gamma(i)}$. Next, we use a parameterized diffeomorphism $f_{\phi, c}: (0,1) \mapsto \mathbb{R}$.  When $\theta$ represents the parameters for a distribution $p_{\theta}$ over the interval $(0,1)$, our model proceeds by either sampling data points as 
$\hat{x}_{i} = f_{\phi, c_{i}}(u_{i}), \ \ u_{i} \sim p_{\theta_{i}}$,
or computing the conditional likelihood:
$p_{\theta_{i}}(f^{-1}_{\phi, c_{i}}(x_{i}))$. Additionally, the decoder's complexity scales as $\mathcal{O}(S(S+H))$, where $S$ represents the number of data points to be inferred and $H$ denotes the number of observed data points.

%It's important to note that both self-attention mechanisms within TACTiS involve pairwise calculations among the variables. Consequently, the encoder's computational complexity scales as $O(N^{2})$, where $N$ is the number of data points in the time series. Conversely, the decoder's complexity scales as $O(S(S+H))$, where $S$ represents the number of data points to be inferred and $H$ denotes the number of observed data points. To address the computational complexity of the encoder in TACTiS, especially for synchronously measured time series, ~\cite{drouin2022tactis} have implemented a temporal transformer variant. This variant applies attention iteratively, first within a time step and then across time steps, effectively reducing the encoder's complexity to $O(n^{2}T + nT^{2})$, where $n$ is the number of variables, and $T$ is approximately the number of time steps. However, it's worth noting that the self-attention-based decoder in TACTiS maintains its complexity scaling as $O(S(S+H))$.

%----------------------------
\begin{figure*}[t]
\centering
\includegraphics[width=.95\textwidth]{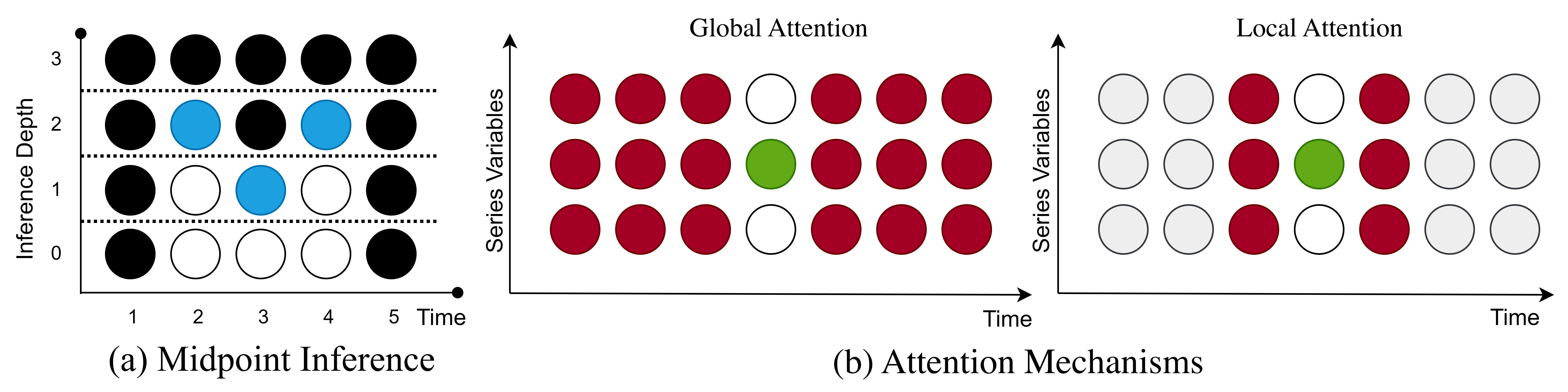}
\caption{(a) Visualization of the midpoint inference mechanism: blue-filled points represent the points earmarked for inference at a particular depth, while black points represent those already observed or inferred at that depth and the white points are unobserved. (b) Comparison between the global attention mechanism and the local attention mechanism, which utilizes a local window containing only the nearest tokens: green-filled points indicate the currently sampled variable, while red  points signify the variables to which the sampled token directs its attention during the sampling process.}
\label{fig:practismidpoint}
\end{figure*}

%-----------------------------
% Theoretical Justification
%-----------------------------
The training procedure of our model involves random permutations to establish an approximately valid copula. Particularly, all margins are approximately the same and uniform. Once a valid copula is obtained, it leverages Sklar's theorem to combine copula density with marginal densities for maximum likelihood estimation (MLE). However,
the conditional copula factorization, as expressed by $c(u_1, \ldots, u_d) = c(u_{\pi_1}) \times c(u_{\pi_2} \mid u_{\pi_1}) \times \cdots \times c(u_{\pi_d} \mid u_{\pi_1}, \ldots, u_{\pi_{d-1}})$ for permutations of indices $\pi$, carries significant implications, especially in the asymptotic limit of infinite permutations. As the order of the marginals becomes irrelevant, the copula converges into a family of exchangeable copulas. 
Relying on exchangeability to ensure the validity of a copula undermines the potential advantages of utilizing a nonparametric copula, ultimately diminishing its expected benefits.

%----------------------------

\subsection{Midpoint Inference for Local Attention}
To enhance computational efficiency while maintaining the prediction performance, we propose the midpoint inference mechanism with temporally local self-attention to effectively reduce computational overhead. Instead of relying on random permutations to establish the conditioning structure, our method employs permutations that recursively infer midpoints within gaps in the observed data. When dealing with a continuous sequence of missing data points for the same variable, we determine the depth of each data point based on the number of midpoint inferences required within that sequence before considering the data point itself as a midpoint. Notably, observed data points are assigned shallower depths compared to data points that are yet to be observed. Consequently, we sample a permutation $\gamma$ that positions data points with shallower assigned depths before those with deeper depths. Here, we determine midpoints by considering the number of data points between the prior observation and the next observation, as visually depicted in Figure~\ref{fig:practismidpoint}. This method is well-suited for regularly or nearly-regularly sampled time series data. 
%It is also straightforward to adapt PrACTiS to define midpoints based on the actual time intervals between data points.
For each data token $\vec{z}_i$, our approach selects a set of conditioning tokens $\vec{H}_i$. These conditioning tokens comprise both past and future windows, consisting of the $k$ closest tokens for each variable in the series that precede $\vec{z}i$ within the generated permutation $\gamma$. Figure~\ref{fig:practismidpoint}(b) illustrates the proposed local-attention conditioning mechanism in comparison with the global self-attention. Here, our model employs learned key and value-generation functions, $K_\mathtt{decode}()$ and $V_\mathtt{decode}()$, to derive distributional parameters $\theta{\gamma(i)}$ for each data point to be inferred, following the ordering imposed by $\gamma$ as follows:
\begin{equation*}
\theta_{i} = \mathcal{A_S} \left (K_\mathtt{decode}(\vec{H}_{i}), \vec{z}_{i}, V_\mathtt{decode}(\vec{H}_{i}) \right )
\end{equation*}

%-----------------------------------
\begin{figure*}[t]
    \centering
    \begin{minipage}[l]{\columnwidth}
        \centering
        \includegraphics[width=.9\columnwidth]{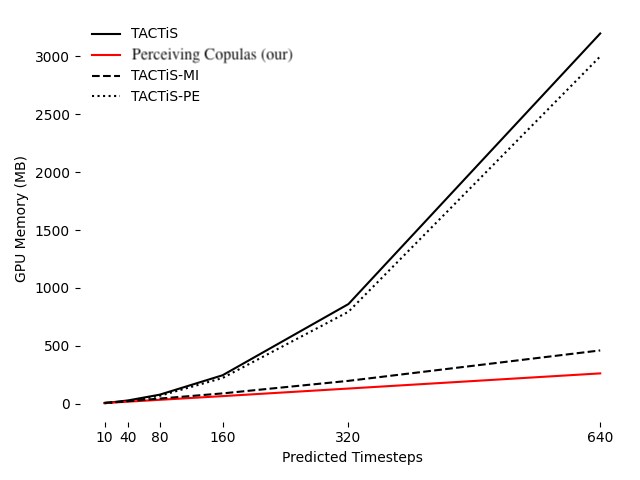}
    \end{minipage}
    \hfill{}
    \begin{minipage}[r]{\columnwidth}
        \centering
        \includegraphics[width=.9\columnwidth]{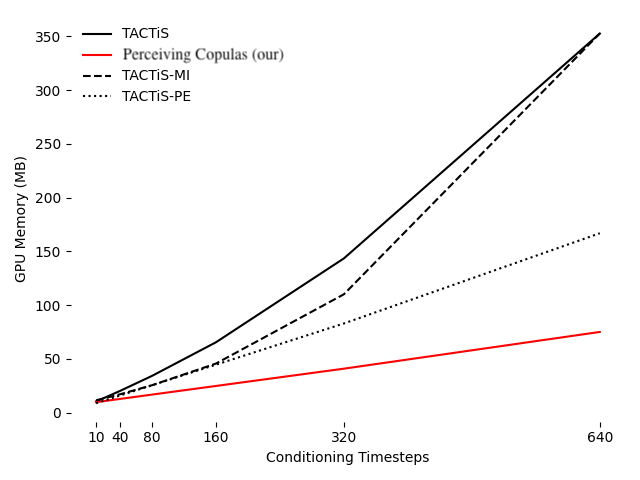}
    \end{minipage}
    \caption{Comparison of memory consumption of perceiver-CDF model (our), TACTiS model, TACTiS model with perceiver-based encoder (TACTiS-PE), and TACTiS model with midpoint imputation (TACTiS-MI) on a synthetic dataset with the varying prediction length (left figure) and the varying conditioning length (right figure).}
    \label{fig:practismem_comp}
\end{figure*}
%-----------------------------------

\subsection{Output Variance Testing}
By incorporating midpoint inference and local attention mechanisms, the decoder adeptly captures dependencies among neighboring imputed samples. However, it is crucial to acknowledge that this enhancement introduces a susceptibility to errors, potentially hindering the training process. To address this concern and prevent error propagation, we propose an output variance testing mechanism for each imputed data point.
In this mechanism, for every imputation, we conduct a series of forecasts by sampling from the derived joint distribution of the missing data. Subsequently, we calculate the output variance of the generated outcomes and compare it with a threshold set to align with the input data variance. If the output variance surpasses four times the threshold, we identify the predicted sample for exclusion in future imputations. In other words, this predicted data point is masked to insulate it from influencing subsequent imputation processes. With a fixed window size, the decoder's complexity can be characterized as $\mathcal{O}(nN)$, where $n$ represents the number of time series variables, and $N$ is the total number of data points in the time series. 

%In summary, the integration of the perceiver-based encoder, the copula-based decoder, the midpoint inference for local attention, and the output variance testing mechanisms significantly enhances the model's performance, leading to notable improvements in both efficiency and scalability.

\begin{table*}[t]
\centering
\caption{Comparison of memory usage and prediction performance between Perceiver-CDF and other approaches in unimodal time series datasets, such as fred-md, traffic, and electricity.}
\label{tab:unimodal-result}
\begin{sc}
\begin{tabular}{l|cccccc}
\hline
\multicolumn{1}{l}{} &\multicolumn{6}{c}{\textbf{\texttt{fred-md - 24 timesteps prediction}}} \\
\multicolumn{1}{l}{\textbf{Approach}} &\multicolumn{1}{c}{\textbf{Params}} &\multicolumn{1}{c}{\textbf{Memory}} &\multicolumn{1}{c}{\textbf{Batches/s}} &\multicolumn{1}{c}{\textbf{NLL}} &\multicolumn{1}{c}{\textbf{RMSE-CM}} &\multicolumn{1}{c}{\textbf{CRPS}} \\
\hline
AR(24)  & 6K   & 3.7 MB  & 37.4 & --                          & $7.0 \scriptstyle \pm 0.5 E+2$  & $1.10 \scriptstyle \pm 0.05$ \\
GPVar   & 78K  & 1.39 GB & 11.7 & $42.3 \scriptstyle \pm 0.6$ & $6.8 \scriptstyle \pm 0.5 E+2$  & $0.86 \scriptstyle \pm 0.06$ \\
TACTiS  & 91K  & 1.51 GB & 12.3 & $42.3 \scriptstyle \pm 0.4$ & $6.1 \scriptstyle \pm 0.4 E+2$  & $0.74 \scriptstyle \pm 0.05$ \\
Perceiver-CDF  & 122K & 1.66 GB & 16.6 & $34.2 \scriptstyle \pm 0.3$   & \boldmath$6.0 \scriptstyle \pm 0.4 E+2$  & \boldmath$0.71 \scriptstyle \pm 0.06$ \\ 
\hline

\hline
\multicolumn{1}{l}{} &\multicolumn{6}{c}{\textbf{\texttt{traffic -  48 timesteps prediction}}} \\
\multicolumn{1}{l}{\textbf{Approach}} &\multicolumn{1}{c}{\textbf{Params}} &\multicolumn{1}{c}{\textbf{Memory}} &\multicolumn{1}{c}{\textbf{Batches/s}} &\multicolumn{1}{c}{\textbf{NLL}} &\multicolumn{1}{c}{\textbf{RMSE-CM}} &\multicolumn{1}{c}{\textbf{CRPS}} \\
\hline
AR(48)  & 20K  & 11.5 MB & 10.31 & --                            & $0.053 \scriptstyle \pm 0.005$  & $0.431  \scriptstyle \pm 0.004$  \\
GPVar   & 78K  & 4.67 GB & 5.81  & $204.6 \scriptstyle \pm 0.8$  & $0.044 \scriptstyle \pm 0.003$  & $0.215 \scriptstyle \pm 0.008$ \\
TACTiS  & 91K  & 5.52 GB & 5.84  & $198.7 \scriptstyle \pm 0.6$  & $0.035 \scriptstyle \pm 0.002$  & $0.181 \scriptstyle \pm 0.009$ \\
Perceiver-CDF  & 122K & 2.75 GB & 5.95  & $188.7 \scriptstyle \pm 0.6$  & \boldmath$0.028 \scriptstyle \pm 0.002$  & \boldmath $0.162 \scriptstyle \pm 0.006$ \\ 
\hline

\multicolumn{1}{l}{} &\multicolumn{6}{c}{\textbf{\texttt{electricity - 48 timesteps prediction}}} \\
\multicolumn{1}{l}{\textbf{Approach}} &\multicolumn{1}{c}{\textbf{Params}} &\multicolumn{1}{c}{\textbf{Memory}} &\multicolumn{1}{c}{\textbf{Batches/s}} &\multicolumn{1}{c}{\textbf{NLL}} &\multicolumn{1}{c}{\textbf{RMSE-CM}} &\multicolumn{1}{c}{\textbf{CRPS}} \\
\hline
AR(48)  & 20K  & 11.6 MB & 10.34 & --                           & $90 \scriptstyle \pm 0.1$ & $0.149 \scriptstyle \pm 0.001$ \\
GPVar   & 78K  & 4.78 GB & 5.76  & $185.6 \scriptstyle \pm 0.5$ & $62 \scriptstyle \pm 0.1$ & $0.060 \scriptstyle \pm 0.001$ \\
TACTiS  & 91K  & 5.42 GB & 5.81  & $182.3 \scriptstyle \pm 0.6$ & $49 \scriptstyle \pm 0.1$ & $0.060 \scriptstyle \pm 0.001$ \\
Perceiver-CDF  & 122K & 2.73 GB & 5.93  & $177.8 \scriptstyle \pm 0.8$   & \boldmath$42 \scriptstyle \pm 0.1$ & \boldmath$0.056 \scriptstyle \pm 0.001$ \\ 
\hline

\multicolumn{1}{l}{} &\multicolumn{6}{c}{\textbf{\texttt{electricity - 672 timesteps prediction}}} \\
\multicolumn{1}{l}{\textbf{Approach}} &\multicolumn{1}{c}{\textbf{Params}} &\multicolumn{1}{c}{\textbf{Memory}} &\multicolumn{1}{c}{\textbf{Batches/s}} &\multicolumn{1}{c}{\textbf{NLL}} &\multicolumn{1}{c}{\textbf{RMSE-CM}} &\multicolumn{1}{c}{\textbf{CRPS}} \\
\hline
AR(672) & 270K & 47.7 MB & 1.74 & --                             & $159 \scriptstyle \pm 0.8$ & $0.290 \scriptstyle \pm 0.02$  \\
GPVar   & 78K  & 4.81 GB & 3.48 & $3.5 \scriptstyle \pm 0.4 E+3$ & $147 \scriptstyle \pm 0.5$ & $0.198 \scriptstyle \pm 0.005$ \\
TACTiS  & 91K  & 4.81 GB & 3.65 & $2.8 \scriptstyle \pm 0.2 E+3$ & $141 \scriptstyle \pm 0.3$ & $0.186 \scriptstyle \pm 0.006$ \\
Perceiver-CDF  & 122K & 372 MB  & 18.3 & $185 \scriptstyle \pm 0.9$       & \boldmath$98 \scriptstyle \pm 0.1$  & \boldmath$0.133 \scriptstyle \pm 0.001$ \\ 
\hline
\end{tabular}
\end{sc}
\end{table*}

%------------------------------------------------------------------------
\section{Experimental Study}
We present comprehensive experiments to showcase the computational efficiency of our proposed perceiver-CDF model. First, we conduct memory consumption scaling experiments using synthetic random walk data to demonstrate the memory efficiency of our proposed model. Next, we evaluate the predictive capabilities of our model, comparing it against the state-of-the-art approaches, such as deep autoregressive AR~\citep{kalliovirta2015gaussian}, GPVar~\citep{salinas2019high}, SSAE-LSTM~\citep{zhu2021multimodal}, and TACTiS~\citep{drouin2022tactis}. Our evaluation spans across three unimodal time series datasets from the Monash Time Series Forecasting Repository~\citep{godahewa2021monash}, including \texttt{electricity}, \texttt{traffic}, and \texttt{fred-md}, for short-term and long-term prediction tasks. Moreover, we evaluate the multi-modality capabilities of our perceiver-based model in three multimodal time series datasets from the UCI Machine Learning Repository~\citep{uci_dataset}, including \texttt{room occupation}~\citep{misc_occupancy_detection__357}, \texttt{interstate traffic}~\citep{misc_metro_interstate_traffic_volume_492}, and \texttt{air quality}~\citep{misc_beijing_multi-site_air-quality_data_501} datasets. The experimental results show the efficacy of our proposed model over other approaches in prediction performance and memory utilization.

\subsection{Memory Consumption Scaling}
In this experiment, we evaluate the computational costs associated with our proposed perceiver-CDF model and the state-of-the-art TACTiS model with respect to the quantity of observed and inferred data. Here, we use the synthetic Random Walk data with a synchronously-measured time series consisting of $10$ variables, $10$ observed time-steps, and $10$ to-be-inferred time-steps. Additionally, we vary the number of observed and inferred time-steps to assess their impact. Our analysis extends to comparing our model with TACTiS-PE, which leverages the perceiver-based encoder architecture for the TACTiS model. Additionally, we consider the TACTiS model with a midpoint inference mechanism, called TACTiS-MI. This model deduces data points using midpoint imputation for temporally local attention.
A comprehensive comparison of memory usage among these models when applied to a single input series is illustrated in Figure~\ref{fig:practismem_comp}. Firstly, it shows the quadratic relationship between the computational cost of TACTiS and the quantity of input data. Secondly, it underscores the remarkable efficiency of the proposed model in terms of memory utilization. Additionally, it showcases the improvements achieved by utilizing perceiver-based encoder and midpoint inference mechanism for TACTiS model. TACTiS-PE, which utilizes a global inference mechanism, operates quadratically when dealing with inferred variables, thereby maintaining its quadratic scaling with respect to the number of predicted time-steps. Conversely, TACTiS-MI employs TACTiS' encoder, preserving its quadratic scaling with respect to the number of observed time steps. Overall, these results underscore the success of the perceiver-CDF model and the proposed midpoint inference mechanism in efficiently mitigating the inherent quadratic scaling issue within TACTiS.

\begin{table*}[t]
\centering
\caption{Comparison of memory usage and prediction performance between Perceiver-CDF and other approaches in multimodal time series datasets, such as room occupation, interstate traffic, air quality.}
\begin{sc}
\label{tab:multimodal-result}
\begin{tabular}{l|ccccc}
\hline
\multicolumn{1}{l}{} &\multicolumn{5}{c}{\textbf{\texttt{room occupation - 6 features attributions}}} \\
\multicolumn{1}{l}{\textbf{Approach}} &\multicolumn{1}{c}{\textbf{Params}} &\multicolumn{1}{c}{\textbf{Memory}} &\multicolumn{1}{c}{\textbf{RMSE-CM}} &\multicolumn{1}{c}{\textbf{Use/No Use}} &\multicolumn{1}{c}{\textbf{High/Low CO$_2$ }}\\
\hline
SSAE-LSTM~ & 76K  & 5.22 GB & $0.056 \scriptstyle \pm 0.002$ & 97.1\% & 96.5\% \\
TACTiS   & 91K  & 6.38 GB & $0.031 \scriptstyle \pm 0.001$ & 98.1\% & 97.7\% \\
Perceiver-CDF    & 122K & 3.09 GB & \boldmath$0.018 \scriptstyle \pm 0.001$ & \boldmath$98.9\%$  & \boldmath$98.4\%$ \\ 
\hline

\hline
\multicolumn{1}{l}{} &\multicolumn{5}{c}{\textbf{\texttt{interstate traffic - 8 features attributions}}} \\
\multicolumn{1}{l}{\textbf{Approach}} &\multicolumn{1}{c}{\textbf{Params}} &\multicolumn{1}{c}{\textbf{Memory}} &\multicolumn{1}{c}{\textbf{RMSE-CM}} &\multicolumn{1}{c}{\textbf{Rain/No Rain}} &\multicolumn{1}{c}{\textbf{High/Low Traffic}}\\
\hline
SSAE-LSTM   & 76K  & 5.68 GB & $0.083 \scriptstyle \pm 0.004$ & 95.3\% & 94.6\% \\
TACTiS  & 91K  & 7.13 GB & $0.065 \scriptstyle \pm 0.003$ & 96.7\% & 96.1\% \\
Perceiver-CDF  & 122K & 3.22 GB & \boldmath$0.027 \scriptstyle \pm 0.003$ & \boldmath$98.2\%$  & \boldmath$97.8\%$ \\ 
\hline

\multicolumn{1}{l}{} &\multicolumn{5}{c}{\textbf{\texttt{air quality - 12 features attributions}}} \\
\multicolumn{1}{l}{\textbf{Approach}} &\multicolumn{1}{c}{\textbf{Params}} &\multicolumn{1}{c}{\textbf{Memory}} &\multicolumn{1}{c}{\textbf{RMSE-CM}} &\multicolumn{1}{c}{\textbf{Rain/No Rain}} &\multicolumn{1}{c}{\textbf{High/Low PM2.5}}\\
\hline
SSAE-LSTM & 76K  & 6.17 GB & $0.106 \scriptstyle \pm 0.006$ & 93.7\% & 93.4\% \\
TACTiS  & 91K  & 8.83 GB & $0.074 \scriptstyle \pm 0.005$ & 95.8\% & 94.9\% \\
Perceiver-CDF  & 122K & 3.41 GB & \boldmath$0.022 \scriptstyle \pm 0.004$ & \boldmath$98.5\%$  & \boldmath$98.1\%$ \\ 
\hline
\end{tabular}
\end{sc}
\label{tab:multimodal}
\end{table*}

\subsection{Forecasting on Unimodal Datasets}
We evaluate our proposed model's computational cost and inference performance across three real-world unimodal datasets. To begin, we employ the \texttt{fred-md} time series dataset, consisting of $20$ input variables, each comprising $24$ observed samples, to predict $24$ time-steps into the future. 
Appendix~\ref{app:performance} provides insights into the models and their training procedures.
Table \ref{tab:unimodal-result} presents a comparative analysis of performance metrics for perceiver-CDF, TACTiS, GPVar, and AR models. We evaluate these models based on negative log-likelihoods (NLL), root-mean-squared-errors of conditional expectations (RMSE-CM), and continuous ranked probability scores (CRPS). 
In Figure~\ref{fig:practisfred}, we demonstrate example inferences generated by our perceiver-CDF and TACTiS model. 
Our proposed model outperforms GPVar and AR while achieving competitive results with TACTiS in both RMSE-CM and CRPS metrics.

Next, we utilize \texttt{traffic} time series data with 20 input variables, each with $48$ observed samples to predict $48$ time-steps ahead. The performance comparisons between our model and other methods are demonstrated in Table \ref{tab:unimodal-result}. 
We also present example inferences from perceiver-CDF and TACTiS in Figure~\ref{fig:practistraffic}. 
Our proposed model demonstrates a significant performance advantage over TACTiS, GPVar, and AR, excelling in both RMSE-CM and CRPS metrics. Notably, we achieve 20\% improvement over TACTiS in terms of RMSE-CM. The number of parameters and memory usage also highlight the efficiency of the proposed model, which utilizes less than 50\% of the memory compared to TACTiS and GPVar.

Lastly, we evaluate our model in the context of short-term and long-term prediction tasks using the \texttt{electricity} dataset. In the short-term prediction experiment, we utilize 20 variables, each spanning $48$ observed time-steps, to forecast $48$ time-steps into the future. As shown in Table \ref{tab:unimodal-result}, our proposed model significantly outperforms other approaches, boasting a 14\% improvement in RMSE-CM compared to TACTiS, all while utilizing just 50\% of available memory. 
Figure~\ref{fig:practiselectricity} illustrates the examples of predictions made by perceiver-CDF and TACTiS, respectively. 
For the long-term prediction task, we work with 10 variables, each encompassing 672 observed time-steps, to predict the subsequent 672 time-steps. This experiment provides valuable insights into the capabilities of these models on a large-scale dataset. Visual representations of the predictions from perceiver-CDF and TACTiS are shown in Figure ~\ref{fig:practiselectricitylong}. In this scenario, our model demonstrates a significant performance advantage over TACTiS, excelling in both RMSE-CM and CRPS while utilizing only 10\% of available memory. Additionally, our perceiver-CDF model manages to capture the seasonal patterns in the data, albeit not as accurately as in the short-term task. Conversely, TACTiS and other methods face inherent challenges when dealing with extended time series. In particular, TACTiS struggles to model the underlying seasonal structures within the data, resulting in less reliable performance when tasked with long-term predictions.

\subsection{Forecasting on Multimodal Datasets}
We first evaluate the predictive capabilities of the perceiver-CDF model on the \texttt{room occupation} dataset~\citep{misc_occupancy_detection__357}. This dataset is multimodal, consisting of $6$ feature attributes related to room conditions, such as temperature, humidity, and CO$_2$ levels. A detailed dataset description is available in Appendix~\ref{app:multimodal}. The evaluation of predictive performance is based on the average RMSE-CM across all six attributes. Furthermore, we undertake two classification tasks: the first task involves predicting room occupancy, while the second task focuses on detecting high CO$_2$ levels (i.e., levels exceeding 700 ppm). Here, we conduct a comparative analysis with TACTiS~\citep{drouin2022tactis} and SSAE-LSTM~\citep{zhu2021multimodal}. Both of these methods employ a strategy of concatenating all feature attributes at each time-step for prediction. The performance results, as presented in Table~\ref{tab:multimodal}, consist of measures such as average RMSE-CM, room occupation detection accuracy, and high CO$_2$ detection accuracy. The memory usage is also provided to highlight the efficiency of our model when achieving 40\% reduction in RMSE-CM compared to TACTiS while utilizing only half of the computational resources. 

Next, we extend our experimentation to the \texttt{interstate traffic} dataset~\citep{misc_metro_interstate_traffic_volume_492}. This dataset comprises $8$ feature attributes related to weather conditions (e.g., temperature, snow), holiday status, and traffic volume. A detailed dataset description is available in Appendix~\ref{app:multimodal}. To assess predictive performance, we utilize RMSE-CM calculated across all eight attributes. Additionally, we investigate two classification tasks: firstly, identifying instances of rainy weather conditions, and secondly, detecting periods of high traffic volume (i.e., volumes exceeding 2000 cars). Table~\ref{tab:multimodal} illustrates that the proposed perceiver-CDF model significantly outperforms other approaches while maintaining linear memory usage. Notably, our approach achieves a 58\% improvement in RMSE-CM compared to TACTiS and consistently excels in prediction tasks related to detecting rain and high traffic. 

Lastly, we evaluate the performance of our approach on the \texttt{air quality} dataset~\citep{misc_beijing_multi-site_air-quality_data_501}, which encompasses $12$ variables, each with $12$ feature attributes, including $6$ pollution-related features (e.g., PM2.5, PM10) and $6$ weather-related features (e.g., temperature, rain). A detailed dataset description is available in Appendix~\ref{app:multimodal}. To assess the quality of our predictions, we employ the average RMSE-CM calculated across all attributes. Moreover, we tackle two classification tasks: firstly, identifying instances of rainy weather conditions, and secondly, detecting periods with elevated PM2.5 levels, specifically those exceeding 80 $\mu g/m^3$. Table~\ref{tab:multimodal} showcases the performance comparison between perceiver-CDF and other approaches, with our model achieving a remarkable 70\% improvement in RMSE-CM compared to TACTiS while utilizing only 40\% of the memory resources.

%The results of our experiments illustrate that the proposed model exhibits increasing efficiency as both the prediction length and the number of feature attributes grow in scale. This efficiency gain becomes particularly pronounced when compared to TACTiS and other existing approaches. In essence, the perceiver-based encoding, the midpoint inference for local inference, and output variance testing mechanisms have proven to be highly effective in addressing the challenges posed by complex multimodal datasets.

\section{Conclusion}
We present a new method for modeling multimodal time series, leveraging cross-attention and copula-attention mechanisms. Our model adeptly encodes the global patterns within partially observed multimodal time series into latent representations, effectively streamlining computational complexity. It also incorporates temporally local attention via midpoint inference, focusing token attention on those with the utmost temporal relevance to their conditioning for precise conditional modeling. 
Our experiments demonstrate that our proposed model exhibits heightened efficiency as prediction length and the number of feature attributes increase. In essence, the combination of perceiver-based encoding, midpoint inference for local context, and output variance testing mechanisms proves highly effective in addressing the challenges posed by intricate multimodal datasets.

\subsection*{Acknowledgments}
We extend our gratitude to Duke University and the Rhodes Family for their generous partial support of this work.  Vahid Tarokh was partially supported by the Office of Naval Research (ONR) under grant number N00014-21-1-2590.

\subsection*{Potential Broader Impact Statement}
This paper presents work whose goal is to advance the field of Machine Learning and time series prediction. There are many potential societal consequences of our work, none of which we feel must be specifically highlighted here.

\nocite{langley00}

\bibliography{ref}
\bibliographystyle{icml2024}

%%%%%%%%%%%%%%%%%%%%%%%%%%%%%%%%%%%%%%%%%%%%%%%%%%%%%%%%%%%%%%%%%%%%%%%%%%%%%%%
%%%%%%%%%%%%%%%%%%%%%%%%%%%%%%%%%%%%%%%%%%%%%%%%%%%%%%%%%%%%%%%%%%%%%%%%%%%%%%%
% APPENDIX
%%%%%%%%%%%%%%%%%%%%%%%%%%%%%%%%%%%%%%%%%%%%%%%%%%%%%%%%%%%%%%%%%%%%%%%%%%%%%%%
%%%%%%%%%%%%%%%%%%%%%%%%%%%%%%%%%%%%%%%%%%%%%%%%%%%%%%%%%%%%%%%%%%%%%%%%%%%%%%%
\newpage
\appendix
\onecolumn

\section{Additional Details for Memory Consumption Experiments}
\label{app:memory_consumption}

To illustrate how the computational demands of the perceiver-CDF and TACTiS models scale in relation to the quantity of observed and inferred data, we investigate the memory usage of these two architectures during training using synthetic random walk data. In this experimental setup, we initiate with a synchronously-measured Random Walk time series comprising $10$ variables, encompassing $10$ observed time-steps and an additional $10$ time-steps to be inferred. We then systematically vary both the number of observed time-steps and the number of inferred time-steps. The model parameters employed for our memory scaling experiments with the perceiver-CDF and TACTiS models are detailed in Table~\ref{tab:memory_scale}. 

Our analysis also extends to comparing the perceiver-CDF model with the TACTiS model that utilizes the perceiver-based encoding mechanism, called TACTiS-PE. It employs similar encoding and decoding mechanisms as TACTiS but leverages the perceiver-based encoder. We also consider the TACTiS model with a midpoint inference mechanism, called TACTiS-MI. This model deduces data points using midpoint imputation and temporally local attention. As shown in Figure~\ref{fig:practismem_comp}, both of these models always outperform the original TACTiS, demonstrating the efficacy of the added mechanism.

\begin{table}[t]
\caption{Model Parameters for Memory Consumption Scaling Experiment.}
\label{tab:memory_scale}
\begin{sc}
\begin{minipage}{.5\textwidth}
  \centering
    %\caption{PrACTiS}
    \begin{tabular}{ll}
        \hline
        \multicolumn{2}{l}{\textbf{(a) Perceiver-CDF Model}} \\ \hline
        \hline
        \multicolumn{2}{l}{Input Encoding} \\ \hline
        Series Embedding Dim. & 5 \\
        Input Encoder Layers & 3 \\ \hline
        \multicolumn{2}{l}{Positional Encoding} \\ \hline
        Dropout & 0.0 \\ \hline
        \multicolumn{2}{l}{Perceiver Encoder} \\ \hline
        Num. Latents & 64 \\
        Latent Dim. & 48 \\
        Attention Layers & 3 \\
        Self-attention Heads & 3 \\
        Cross-attention Heads & 3 \\
        Dropout & 0.0 \\ \hline
        \multicolumn{2}{l}{Perceiver Decoder} \\ \hline
        Cross-attention Heads & 3 \\ \hline
        \multicolumn{2}{l}{Copula Decoder} \\ \hline
        Min. u & 0.01 \\
        Max. u & 0.99 \\ \hline
        \multicolumn{2}{l}{Attentional Copula} \\ \hline
        Attention Layers & 3 \\
        Attention Heads & 3 \\
        Attention Dim. & 16 \\
        Feedforward Dim. & 16 \\
        Feedforward Layers & 3 \\
        Resolution & 50 \\ \hline
        \multicolumn{2}{l}{Marginal Flow} \\ \hline
        Feedforward Layers & 2 \\
        Feedforward Dim. & 8 \\
        Flow Layers & 2 \\
        Flow Dim. & 8 \\ 
        \hline
    \end{tabular}
\end{minipage}
\hfill
\begin{minipage}{.5\textwidth}
    \centering
    %\caption{TACTiS}
    \begin{tabular}{ll}
        \hline
        \multicolumn{2}{l}{\textbf{(b) TACTiS Model}} \\ \hline
        \hline
        \multicolumn{2}{l}{Input Encoding} \\ \hline
        Series Embedding Dim. & 5 \\
        Input Encoder Layers & 3 \\ \hline
        \multicolumn{2}{l}{Positional Encoding} \\ \hline
        Dropout & 0.0 \\ \hline
        \multicolumn{2}{l}{Temporal Encoder} \\ \hline
        Attention Layers & 3 \\
        Attention Heads & 3 \\
        Attention Dim. & 16 \\
        Attention Feedforward Dim. & 16 \\
        Dropout & 0.0 \\ \hline
        \multicolumn{2}{l}{Copula Decoder} \\ \hline
        Min. u & 0.01 \\
        Max. u & 0.99 \\ \hline
        \multicolumn{2}{l}{Attentional Copula} \\ \hline
        Attention Layers & 3 \\
        Attention Heads & 3 \\
        Attention Dim. & 16 \\
        Feedforward Dim. & 16 \\
        Feedforward Layers & 3 \\
        Resolution & 50 \\ \hline
        \multicolumn{2}{l}{Marginal Flow} \\ \hline
        Feedforward Layers & 2 \\
        Feedforward Dim. & 8 \\
        Flow Layers & 2 \\
        Flow Dim. & 8 \\ 
        \hline
    \end{tabular}
\end{minipage}
\end{sc}
\end{table}

\section{Additional Details for Inference Performance Experiments}
\label{app:performance}
The TACTiS model parameters employed for these experiments were adopted from the configuration used by ~\cite{drouin2022tactis}. We also adopt these parameters as the foundation for establishing the comparable perceiver-CDF model. Additionally, we employ deep AR(d) models, characterized as feedforward models that take the previous $d$ time-steps as inputs to predict the current time-step. Below, in Table~\ref{tab:performance}, we provide a comprehensive listing of the model parameters utilized for our perceiver-CDF and TACTiS models.

\begin{table}[t]
\caption{Model Parameters for Performance Experiments.}
\label{tab:performance}
\begin{sc}
\begin{minipage}{.5\textwidth}
    \centering
    %\caption{PrACTiS}
    \begin{tabular}{ll}
        \hline
        \multicolumn{2}{l}{\textbf{(a) Perceiver-CDF Model}} \\ \hline
        \hline
        \multicolumn{2}{l}{Input Encoding} \\ \hline
        Series Embedding Dim. & 5 \\
        Input Encoder Layers & 3 \\ \hline
        \multicolumn{2}{l}{Positional Encoding} \\ \hline
        Dropout & 0.01 \\ \hline
        \multicolumn{2}{l}{Perceiver Encoder} \\ \hline
        Num. Latents & 256 \\
        Latent Dim. & 48 \\
        Attention Layers & 2 \\
        Self-attention Heads & 3 \\
        Cross-attention Heads & 3 \\
        Dropout & 0.01 \\ \hline
        \multicolumn{2}{l}{Perceiver Decoder} \\ \hline
        Cross-attention Heads & 3 \\ \hline
        \multicolumn{2}{l}{Copula Decoder} \\ \hline
        Min. u & 0.05 \\
        Max. u & 0.95 \\ \hline
        \multicolumn{2}{l}{Attentional Copula} \\ \hline
        Attention Layers & 1 \\
        Attention Heads & 3 \\
        Attention Dim. & 16 \\
        Feedforward Dim. & 48 \\
        Feedforward Layers & 1 \\
        Resolution & 20 \\ \hline
        \multicolumn{2}{l}{Marginal Flow} \\ \hline
        Feedforward Layers & 1 \\
        Feedforward Dim. & 48 \\
        Flow Layers & 3 \\
        Flow Dim. & 16 \\ \hline
    \end{tabular}
\end{minipage}
\hfill
\begin{minipage}{.5\textwidth}
    \centering
    %\caption{TACTiS}
    \begin{tabular}{ll}
        \hline
        \multicolumn{2}{l}{\textbf{(b) TACTiS Model}} \\ \hline
        \hline
        \multicolumn{2}{l}{Input Encoding} \\ \hline
        Series Embedding Dim. & 5 \\
        Input Encoder Layers & 3 \\ \hline
        \multicolumn{2}{l}{Positional Encoding} \\ \hline
        Dropout & 0.01 \\ \hline
        \multicolumn{2}{l}{Temporal Encoder} \\ \hline
        Attention Layers & 2 \\
        Attention Heads & 2 \\
        Attention Dim. & 24 \\
        Attention Feedforward Dim. & 24 \\
        Dropout & 0.01 \\ \hline
        \multicolumn{2}{l}{Copula Decoder} \\ \hline
        Min. u & 0.05 \\
        Max. u & 0.95 \\ \hline
        \multicolumn{2}{l}{Attentional Copula} \\ \hline
        Attention Layers & 1 \\
        Attention Heads & 3 \\
        Attention Dim. & 16 \\
        Feedforward Dim. & 48 \\
        Feedforward Layers & 1 \\
        Resolution & 20 \\ \hline
        \multicolumn{2}{l}{Marginal Flow} \\ \hline
        Feedforward Layers & 1 \\
        Feedforward Dim. & 48 \\
        Flow Layers & 3 \\
        Flow Dim. & 16 \\ \hline
    \end{tabular}
\end{minipage}
\end{sc}
\end{table}

\subsection{Short-term Forecasting Experiments}
The datasets, namely \texttt{fred-md}, \texttt{traffic}, and \texttt{electricity}, all consist of 20 input variables. Our model training process for these datasets follows a consistent protocol: we use batch sizes of $24$ for the \texttt{fred-md} dataset and $48$ for the \texttt{traffic} and \texttt{electricity} datasets. This training routine extends over $100$ epochs, each comprising $512$ batches. For optimizing all our models, we employ the RMSProp optimizer \citep{hinton2012neural} with an initial learning rate set at $1e-3$.

\subsection{Long-term Forecasting Experiments}
Given the considerable computational demands associated with forecasting 672 time-steps in this experiment, we made the strategic decision to reduce the batch size to 1 to facilitate the training of the TACTiS model. However, this adjustment presented challenges when applying perceiver-CDF's midpoint inference scheme, as its initial forecasts extended across hundreds of time steps, leading to suboptimal performance. To overcome this limitation and achieve substantially improved results in the realm of long-term forecasting, we refined the midpoint inference scheme. This refinement involved introducing greater autoregressive behavior at the outset, with a carefully designed sampling order that ensured each time-step remained within a user-defined maximum interval from the nearest conditioning variable. In this experiment, we set a relatively aggressive maximum interval of three time-steps, and the subsequent results reflect the impact of this adjustment.

It's important to highlight that the autoregressive refinement made in the midpoint inference order doe not alter the computational complexity of training the perceiver-CDF model. While this adjustment has the potential to improve forecasting accuracy in specific situations, it's worth acknowledging that it can also pose challenges when predicting time-steps that fall between observed data points.

% fred-md short forecast
%-----------------------------------
\begin{figure}[t]
    \centering
    \begin{minipage}[l]{.5\textwidth}
        \centering
        \includegraphics[width=\textwidth]{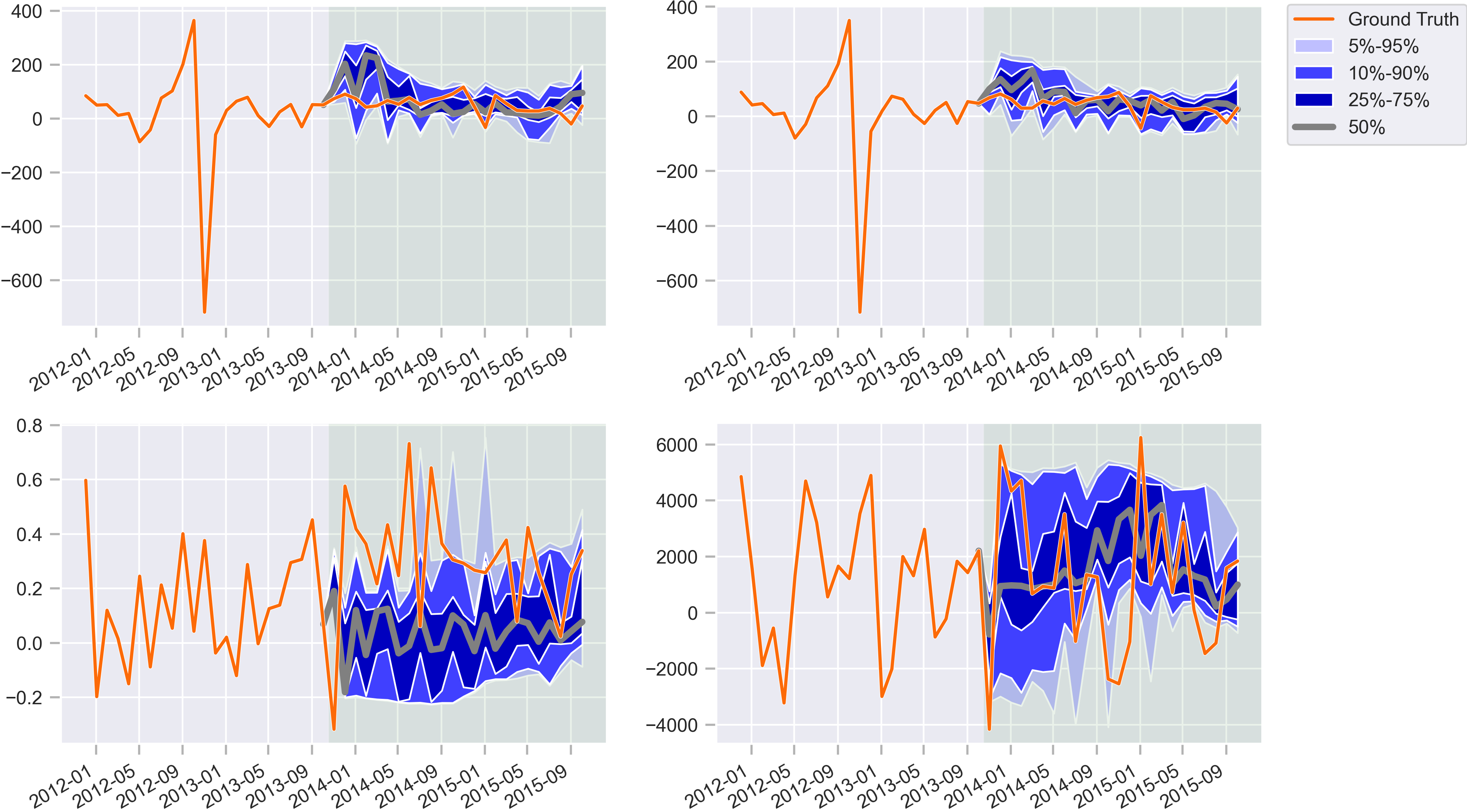}
    \end{minipage}
    %\hfill{}
    \begin{minipage}[r]{.49\textwidth}
        \centering
        \includegraphics[width=\textwidth]{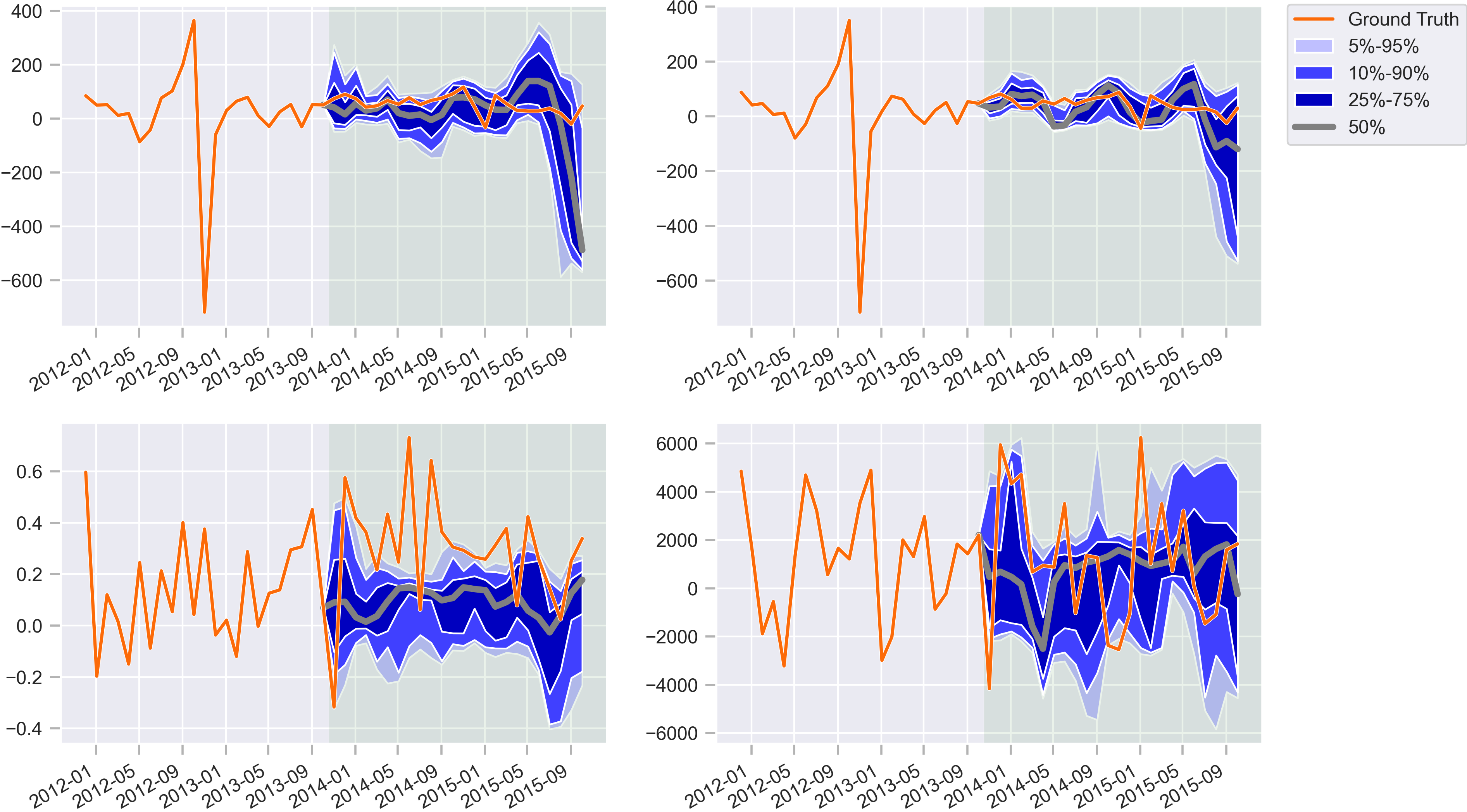}
    \end{minipage}
    \caption{The predicted samples by the perceiver-CDF (left) and TACTiS (right) for two-year forecasts, corresponding to $24$ time-steps, conditioned on two-year historical data in \texttt{fred-md} dataset.}
    \label{fig:practisfred}
\end{figure}

% traffic short forecast
%-----------------------------------
\begin{figure}[t]
    \centering
    \begin{minipage}[l]{.5\textwidth}
        \centering
        \includegraphics[width=\textwidth]{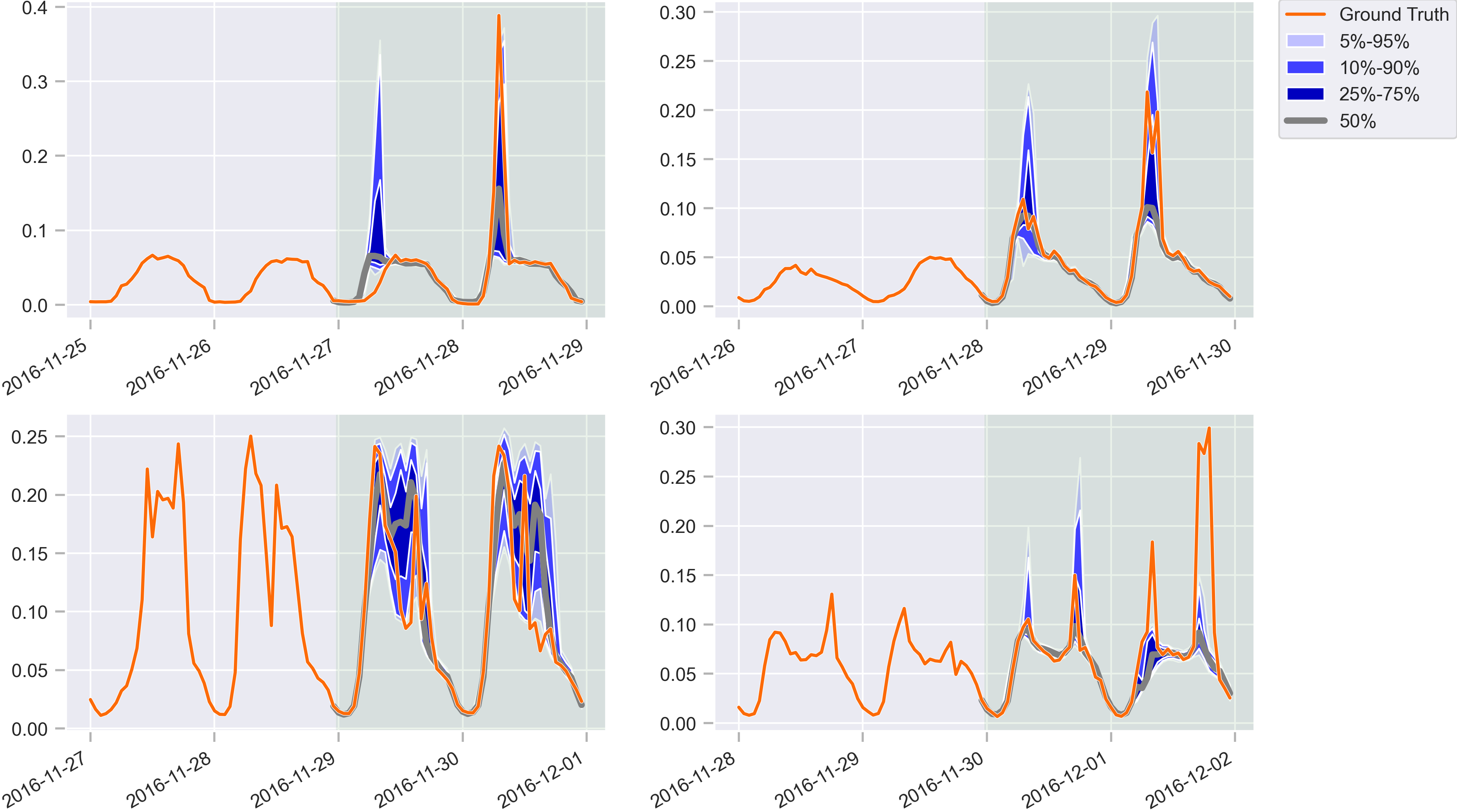}
    \end{minipage}
    %\hfill{}
    \begin{minipage}[r]{.49\textwidth}
        \centering
        \includegraphics[width=\textwidth]{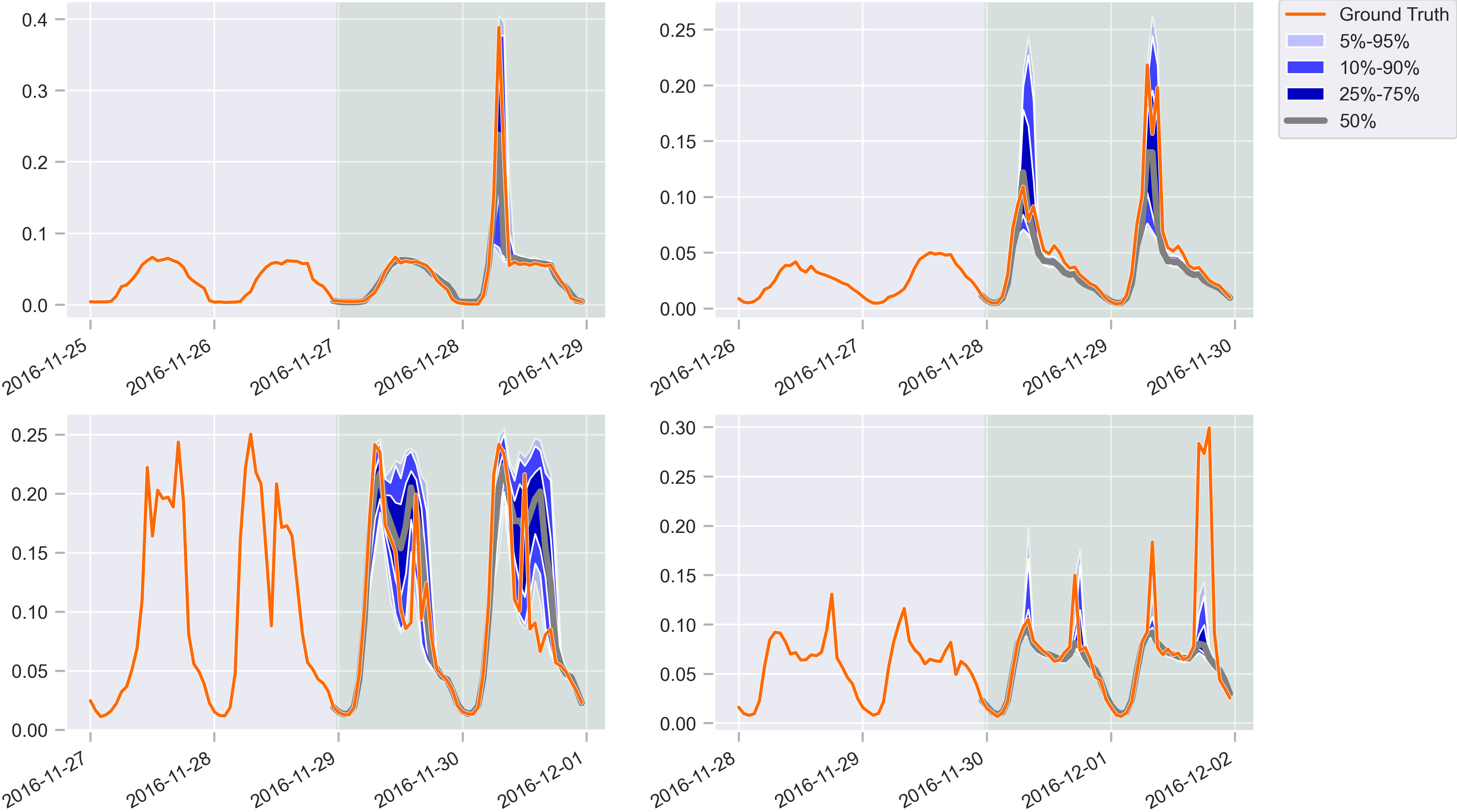}
    \end{minipage}
    \caption{The predicted samples by the perceiver-CDF (left) and TACTiS (right) for two-day forecasts, corresponding to $48$ time-steps, conditioned on two-day historical data in \texttt{traffic} dataset.}
    \label{fig:practistraffic}
\end{figure}

\section{Additional Details for Multimodal Experiments}
\label{app:multimodal}

The \texttt{room occupation} dataset~\citep{misc_occupancy_detection__357} is a comprehensive multimodal dataset encompassing six distinct feature attributes that capture room conditions and room occupancy status (i.e., the primary output). These attributes include temperature, relative humidity, humidity ratio, light levels, and CO$_2$ concentrations. In this experiment, our model is trained to forecast $48$ time-steps ahead, utilizing historical data spanning the preceding $48$ time-steps. The evaluation of predictive performance is based on the average RMSE-CM across all six attributes. Furthermore, we undertake two classification tasks: the first task involves predicting room occupancy, while the second task focuses on detecting high CO$_2$ levels (i.e., levels exceeding 700 ppm).

The \texttt{interstate traffic} dataset~\citep{misc_metro_interstate_traffic_volume_492} presents a collection of multimodal traffic data samples, encompassing eight feature attributes that capture a wide range of information. These attributes are associated with diverse aspects, including weather conditions, temporal factors, holiday status, and traffic volume (i.e., primary output). Here, the weather-related attributes include temperature, precipitation (rain and snow), cloud cover, and the categorization of weather conditions. In this experimental setup, our model is rigorously trained to predict traffic conditions up to $48$ time-steps into the future, leveraging historical data spanning the preceding $48$ time-steps. To assess predictive performance, we utilize RMSE-CM calculated across all eight attributes. Additionally, we investigate two classification tasks: firstly, identifying instances of rainy weather conditions, and secondly, detecting periods of high traffic volume (i.e., volumes exceeding 2000 cars).

The \texttt{air quality}~\citep{misc_beijing_multi-site_air-quality_data_501} dataset comprises pollution measures and weather-related metrics data. It encompasses $12$ variables, each with $12$ feature attributes, including $6$ pollution-related features and $6$ weather-related features. The pollution-related features include PM2.5, PM10, SO$_2$, NO$_2$, CO, O$_3$ concentrations. The weather-related features consist of temperature, dew point temperature, pressure, precipitation, wind direction and speed. Similarly, our model is trained to forecast 48 time-steps into the future, leveraging historical data spanning the preceding 48 time-steps. To assess the quality of our predictions, we employ the average RMSE-CM calculated across all attributes. Moreover, we tackle two classification tasks: firstly, identifying instances of rainy weather conditions, and secondly, detecting periods with elevated PM2.5 levels, specifically those exceeding 80 $\mu g/m^3$. The detailed results of these experiment can be found in Table~\ref{tab:multimodal-result}.

% electric short forecast
%-----------------------------------
\begin{figure}[t]
    \centering
    \begin{minipage}[l]{.5\textwidth}
        \centering
        \includegraphics[width=\textwidth]{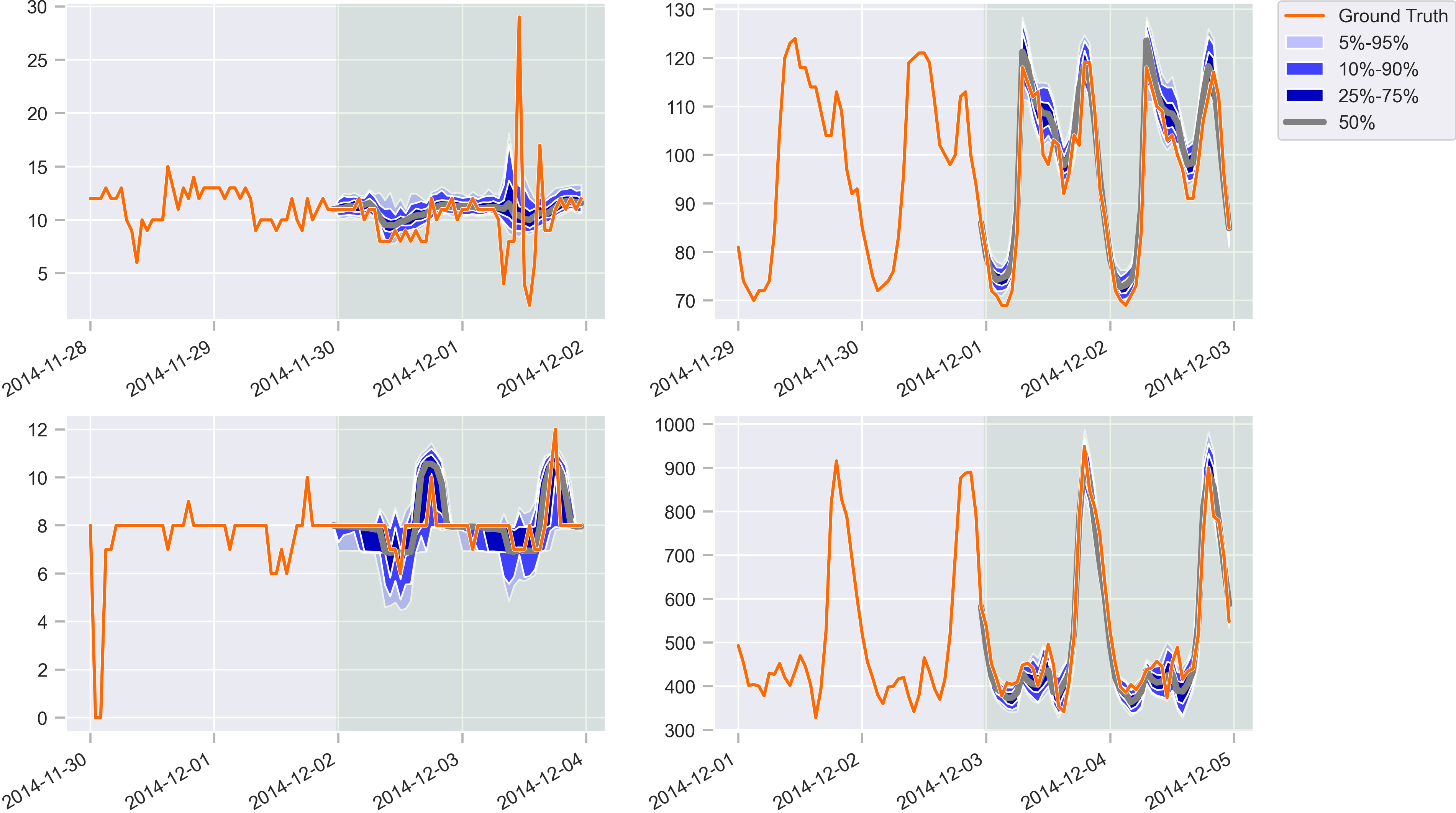}
    \end{minipage}
    %\hfill{}
    \begin{minipage}[r]{.49\textwidth}
        \centering
        \includegraphics[width=\textwidth]{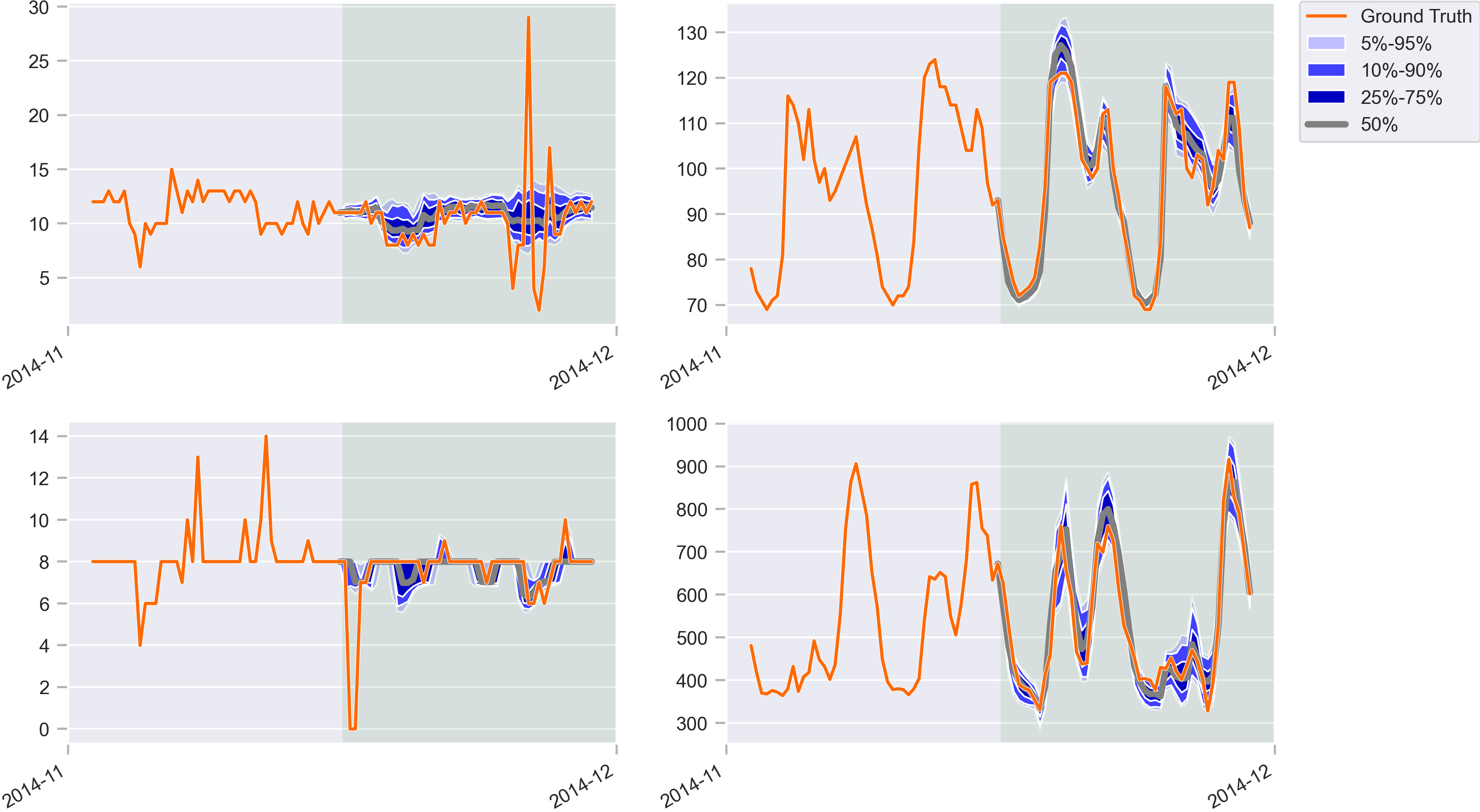}
    \end{minipage}
    \caption{The predicted samples by the perceiver-CDF (left) and TACTiS (right) for two-day forecasts, corresponding to $48$ time-steps, conditioned on two-day historical data in \texttt{electricity} dataset.}
    \label{fig:practiselectricity}
\end{figure}

% electric long forecast
%-----------------------------------
\begin{figure}[t]
    \centering
    \begin{minipage}[l]{.5\textwidth}
        \centering
        \includegraphics[width=\textwidth]{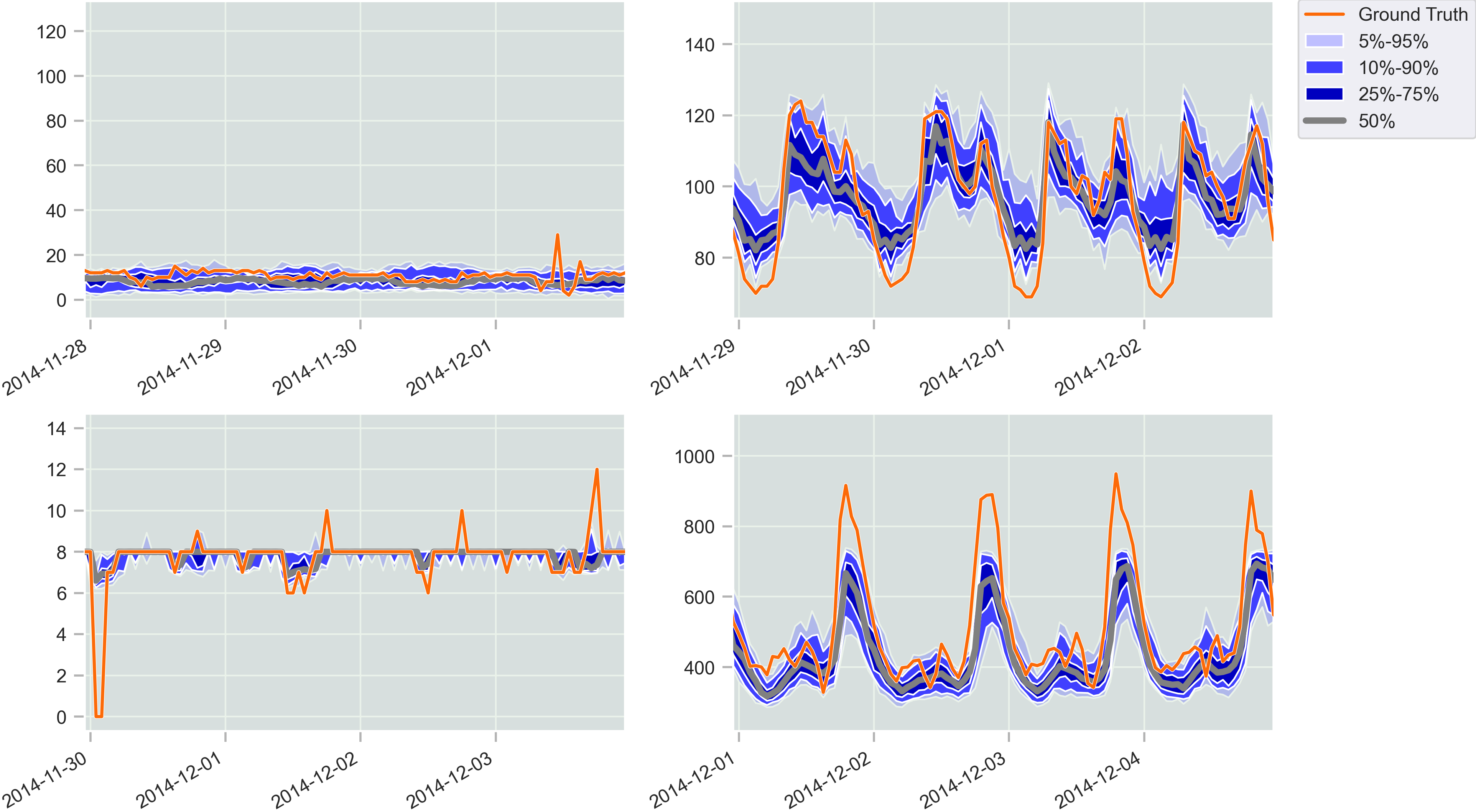}
    \end{minipage}
    %\hfill{}
    \begin{minipage}[r]{.49\textwidth}
        \centering
        \includegraphics[width=\textwidth]{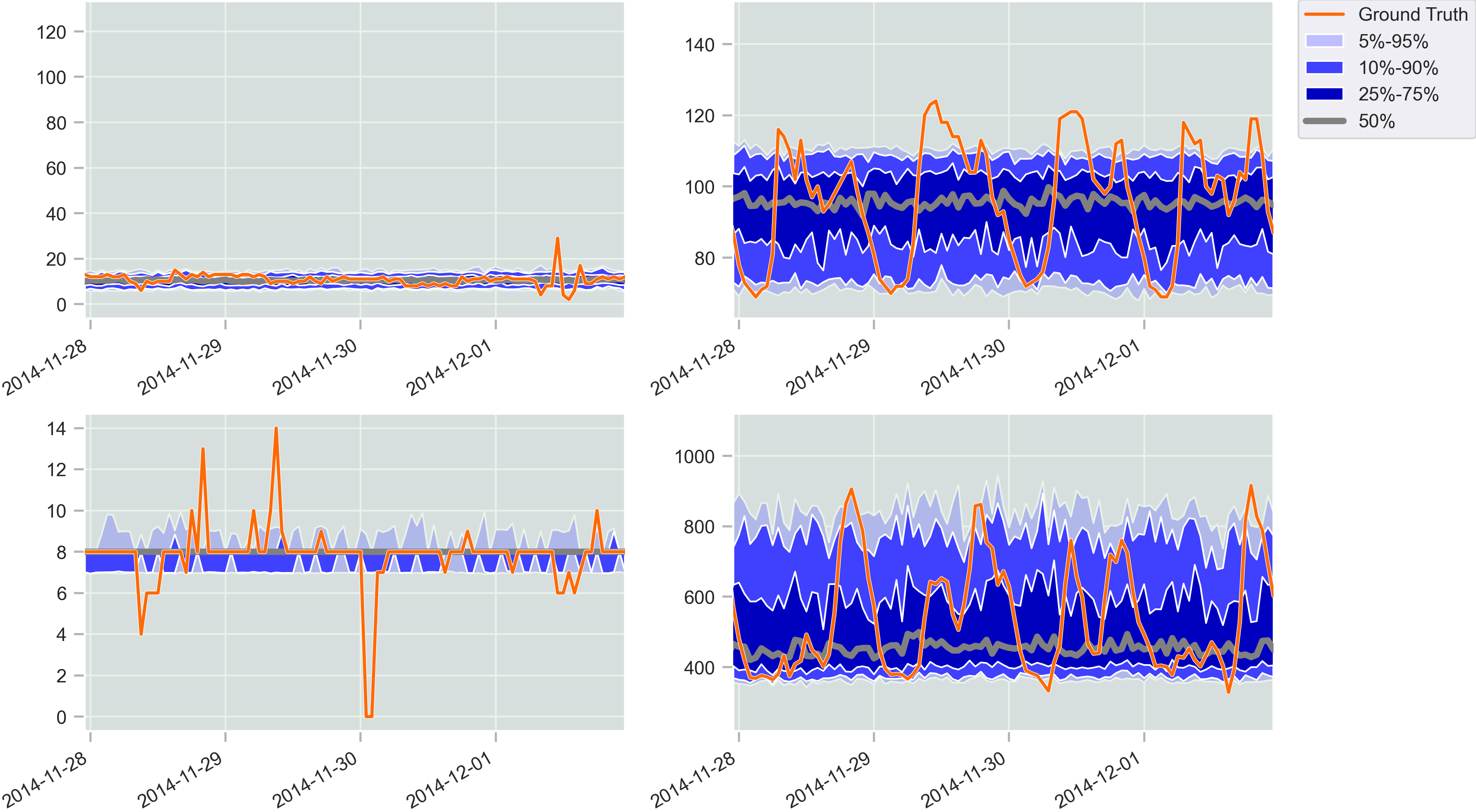}
    \end{minipage}
    \caption{The predicted samples by the perceiver-CDF (left) and TACTiS (right) for one-month forecasts, corresponding to $672$ time-steps, conditioned on one-month historical data in \texttt{electricity} dataset.}
    \label{fig:practiselectricitylong}
\end{figure}

\section{Forcasting Samples}
\label{app:sample}
Here, we provide some predicted samples produced using the perceiver-CDF and TACTiS models. First, Figure~\ref{fig:practisfred} illustrates the predicted samples from the short-term task (i.e., $24$ time-steps, which corresponds to $2$ years) in \texttt{fred-md} dataset using the perceiver-CDF and TACTiS models. Next, Figure~\ref{fig:practistraffic} demonstrates the short-term predicted samples (i.e., $48$ time-step, corresponding to $2$ days) in \texttt{traffic} dataset using the perceiver-CDF and TACTiS models. Similarly, the short-term predicted samples (i.e., $48$ time-step, corresponding to $2$ days) in \texttt{electricity} dataset using the perceiver-CDF and TACTiS models are shown in Figure~\ref{fig:practiselectricity}. Lastly, Figure~\ref{fig:practiselectricitylong} illustrates the last 4-day predicted samples from the long-term task (i.e., $672$ time-step, corresponding to $1$ month) in \texttt{electricity} dataset using the perceiver-CDF and TACTiS models.

\section{Advantages and Disadvantages}
Throughout extensive experiments on unimodal and multimodal datasets, we have observed that the proposed perceiver-CDF model effectively learns to model time series data and contributes to improved prediction performance. Nevertheless, it is worth noting that in certain instances, the utilization of perceiver-CDF may result in a significant increase in memory consumption without proportional gains in performance. For instance, for shorter time series, such as the \texttt{fred-md} dataset in the Monash Time Series Forecasting Repository~\citep{godahewa2021monash}, the perceiver-CDF model may require increased time and computational resources for training compared to other approaches. This heightened resource requirement stems from the overhead associated with the cross-attention mechanism, which is designed to map input embeddings into latent embeddings. In this experiment, our model obtains insignificant improvement over TACTiS, but utilizes more memory resources. 

In contrast, when dealing with long and complex time series, such as those present in the Monash Time Series Forecasting Repository~\citep{godahewa2021monash} (e.g., \texttt{traffic}, \texttt{electricity} datasets), and the UCI Machine Learning Repository~\citep{uci_dataset} (\texttt{room occupation} dataset~\citep{misc_occupancy_detection__357}, \texttt{interstate traffic}~\citep{misc_metro_interstate_traffic_volume_492}, and \texttt{air quality}~\citep{misc_beijing_multi-site_air-quality_data_501} datasets), our experiments clearly demonstrate the advantages of the perceiver-CDF model's enhanced scalability, as exemplified by the outcomes of our memory scaling experiments. These advancements highlight our models' ability to train efficiently on hardware configurations with more modest computational resource allocations, rendering them highly suitable for a wide range of time series modeling tasks. Furthermore, our proposed model exhibits the capacity to handle more complex tasks with a linear increase in memory usage, allowing for greater flexibility in model design. Here, the proposed model simplifies the modeling of time series data that would otherwise pose computational challenges when employing TACTiS or alternative methodologies. As a result, the perceiver-CDF model can adeptly learn to model time series data while consuming only a fraction of the memory resources, all while achieving highly competitive performance levels.
Overall, the perceiver-CDF model shows potential applications in the generative models for handling missing data. Its utilization of latent variables to deduce a global state and enhance computational efficiency bears a striking resemblance to how other generative models leverage their latent variables for efficient missing data inference. The fusion of this latent space with an attentional generative mechanism empowers the perceiver-CDF model to overcome many of the inherent architectural challenges encountered when applying existing generative models to missing data tasks. Attentional models, which replace complex architectural details with information encoded within their input data, may prove to be ideally suited for addressing missing data challenges.

%%%%%%%%%%%%%%%%%%%%%%%%%%%%%%%%%%%%%%%%%%%%%%%%%%%%%%%%%%%%%%%%%%%%%%%%%%%%%%%
%%%%%%%%%%%%%%%%%%%%%%%%%%%%%%%%%%%%%%%%%%%%%%%%%%%%%%%%%%%%%%%%%%%%%%%%%%%%%%%

\end{document}